\documentclass{iosart2c}

\usepackage[T1]{fontenc}
\usepackage{times}%

\usepackage{natbib}
\usepackage{amsmath}
\usepackage{dcolumn}
\usepackage{graphics}

\usepackage{soul}
\usepackage{graphicx}

\newcolumntype{d}[1]{D{.}{.}{#1}}

\firstpage{1} \lastpage{5} \volume{1} \pubyear{2024}

\begin{document}
\begin{frontmatter}                           

%
\title{A multitask learning framework for leveraging subjectivity of annotators to identify misogyny}

\runningtitle{leveraging subjectivity to identify misogyny}

\author[A]{\fnms{Jason} \snm{Angel}},
\author[B]{\fnms{Segun Taofeek} \snm{Aroyehun}},
\author[A]{\fnms{Grigori} \snm{Sidorov}},
and
\author[A]{\fnms{Alexander} \snm{Gelbukh}\thanks{Corresponding author. Phone: (+52) 55 5729 6000 ext. 56544. E-mail: gelbukh@cic.ipn.mx. © Gelbukh, 2024. The definitive, peer reviewed and edited version of this article is published in Journal of Intelligent \& Fuzzy Systems, IOS Press, 2024, DOI:10.3233/JIFS-219355
}}

\runningauthor{Angel et al.}

\address[A]{Instituto Politécnico Nacional (IPN), Centro de Investigación en Computación (CIC), Mexico City, Mexico\\
E-mail: ajason08@gmail.com, \{sidorov, gelbukh\}@cic.ipn.mx}

\address[B]{University of Konstanz, Konstanz, Germany\\
E-mail: aroyehun.segun@gmail.com}

\begin{abstract}
Identifying misogyny using artificial intelligence is a form of combating online toxicity against women. However, the subjective nature of interpreting misogyny poses a significant challenge to model the phenomenon. In this paper, we propose a multitask learning approach that leverages the subjectivity of this task to enhance the performance of the misogyny identification systems. We incorporated diverse perspectives from annotators in our model design, considering gender and age across six profile groups, and conducted extensive experiments and error analysis using two language models to validate our four alternative designs of the multitask learning technique to identify misogynistic content in English tweets. The results demonstrate that incorporating various viewpoints enhances the language models' ability to interpret different forms of misogyny. This research advances content moderation and highlights the importance of embracing diverse perspectives to build effective online moderation systems.
\end{abstract}


\end{frontmatter}

\section{Introduction}

Misogyny is etymologically defined as "hate towards women" and is often expressed as a form of sexism, hate speech, and abuse against women or girls in online and offline environments \cite{misogynymed2017, frenda2019online, sexistslur2020}. 
In the landscape of content moderation, addressing misogyny has emerged as a critical concern, and over the years several artificial intelligence (AI) systems have been proposed to detect and discourage online aggression toward women \cite{mishra2019tackling, istaiteh2020racist}. 
However, the reliability of such initiatives is at least arguable in the light that humans disagree on what should be considered misogynistic. 
Moreover, the lack of a clear source of truth raises ethical concerns about who gets to define misogyny and whether the perspectives included adequately represent marginalized groups who may be most affected by online misogyny.
For instance, because of their lack of knowledge of the phenomenon \cite{waseem2016you, guest2021expert}. The interpretation of misogyny may be also affected by the culture, personal beliefs, and exposition of annotators to different forms of misogyny. Considering this subjectivity is primordial to designing systems to combat misogyny, otherwise, the moderation platform may face negative consequences such as: 
\begin{itemize}
    \item \textbf{Bias and unfairness} in which the system may inadvertently favor certain perspectives, leading to biased or unfair content moderation, resulting in censorship or the failure to address misogyny effectively, depending on whose viewpoint prevails.
    \item \textbf{Reduced user trust} in the application because it inconsistently identifies misogyny either by allowing misogynistic content to persist unchecked or conversely, leading to over-moderation, constraining free expression of users whose not misogynistic content gets flagged.
\end{itemize}

This study proposes four approaches to effectively incorporate the viewpoints of six different profile groups to enhance the detection of misogyny. These profiles are defined by using the gender and age of annotators. We conduct experiments by applying multitask learning techniques in two transformer-based models. We then compare the multitask learning models with their corresponding architecture that uses single-task learning. The single-task learning approaches serve as our baselines. As a vital part of our proposal the reader is welcome to read more about the multitask learning technique in \cite{caruana1997multitask}. In short, the technique employs shared representations to learn and leverage related tasks thereby enhancing the performance of these tasks simultaneously.

The present research provides groundbreaking insights into content moderation by acknowledging that human subjectivity in data annotation have a substantial positive effect on combating misogyny. Our primary goal is to reduce bias in AI systems, fostering fairness and equity in technology. This approach enhances model performance and adaptability to meet various user groups' diverse expectations and contexts. By addressing subjectivity, this work contributes to the development of ethically sound AI systems, aligning technology with societal values through bias mitigation in the prediction of machine learning models for misogyny identification.


\section{Related works}
In this study, our primary focus centers on the challenging task of misogyny identification, a task inherently imbued with subjectivity and often characterized by disagreements among annotators during the labeling process. These disagreements are pertinent, serving as a human model of the phenomenon that machine learning systems endeavor to grasp to accurately identify misogyny. Traditionally, addressing these discrepancies involved aggregating the varied perspectives of annotators into a hard label, frequently determined by a majority vote \cite{annotatingmisogyny2021}.

However, a growing body of research recognizes the relevance of the inherent subjectivity to better design machine learning systems~\cite{lewidi2021, dealingdesagreements2022}. In real-world scenarios, human annotators often hold differing views. In such cases, a soft metric is often introduced to gauge the alignment between the model's predicted label probabilities and the distribution of labels assigned by human annotators.
This approach, particularly in the context of misogyny identification, has led to pioneering projects such as \cite{arabiccorpus2022}, which considers demographic traits and culture in misogyny detection for the Arabic language.

Turning to specific methodologies, notable contributions come from the winners of Task 1 in the Exist2023 shared task initiative \cite{plaza2023overview}. One group employed data augmentation, enriching their model with additional contexts from past competitions \cite{exist2023winner2}. Another team augmented model features with annotators' demographic information \cite{exist2023winner1}. Finally, a distinct approach treated the task as a regression problem, incorporating contrastive learning techniques, both with and without updating learned parameters, into the traditional fine-tuning process \cite{exist2023winner3}.

Among systems utilizing multitask learning (MTL) techniques for sexism classification, \cite{MTLsexism1} explored various auxiliary tasks. These tasks include homogeneous tasks derived through unsupervised learning and weak labeling and heterogeneous tasks involving sarcasm detection and emotion classification. The heterogeneous tasks provide external cues for classification.
Another set of works adopts a hierarchical framework for sexism classification. The initial layer performss binary classification, and subsequent layers classify sexist content into subclasses. This approach aims to improve predictions at each subtask levels.
For instance, in \cite{MTLsexism2}, the authors employed the annotation schema of the EDOS dataset \cite{edosdataset}, where classes represented the reasons behind content being sexist. Similarly, the authors of \cite{MTLexist} leveraged the annotation schema of Exist2023, where subsequent layers represented the intention and categorization of sexist text.

In summary, while traditional methods aggregate perspectives into a hard label, recent works have made an effort to recognize subjectivity in real-world scenarios. 
Our research stands out within the field of misogyny identification as it engages with diverse auxiliary tasks and employs a hierarchical framework. This approach contributes to a comprehensive exploration of misogyny identification and offers insights that extend to both theoretical and practical contexts.

\section{Methodology}
We present an overview of the dataset employed for misogyny identification, outlining the process used to define the profile groups. We then elaborate on the specifications of our models, designed to incorporate a range of annotator perspectives. Then, we describe the methods used to conduct and validate our experiments, providing a comprehensive understanding of our research approach.

\begin{figure*}[t!]
\label{f1}
\includegraphics[width=12cm]{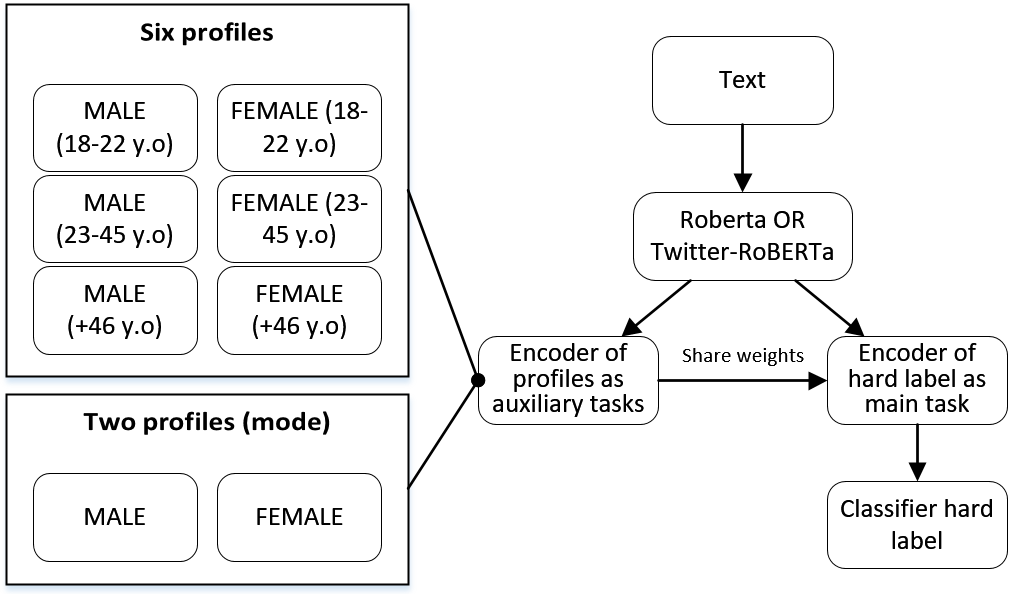}
\caption{General architecture design for incorporating the annotator profile perspectives as auxiliary tasks to support the learning of the hard label as the main task}
\end{figure*}

\subsection{Dataset}
We utilized the English slice of Exist2023 \cite{plaza2023overview}, this dataset was curated from over 8 million tweets, containing nearly 4.7K English tweets as samples that potentially include abuse toward women. This selection aimed to mitigate potential biases arising from terminology, publication time, or author socio-demographic profiles.

The dataset segment employed was categorized as either sexist or not sexist by 398 annotators, who were diligently segmented by gender (MALE, FEMALE) and age groups (18-22, 23-45, and 46+). Each sample was annotated by one member from each of the six segments, forming the profile groups for this study. To determine the hard label, we simply selected the most frequently occurring label (the mode) and excluded samples where there was a tie between labels. It's important to note that we did not conduct any preprocessing on the textual contents of the dataset.

\subsection{Experimentation setting}

In our experimental setup, we compare the single-task learning approaches as the baselines with the multi-task learning approaches as the intervention models in the task of predicting misogynistic English content. The models are evaluated by their performance in the hard label, which is defined as the majority vote of the six annotators. The baselines rely on the hard label for training and predicting misogynistic content, while the intervention models learn the individual perspective of annotators to support the training and prediction of the hard label. 

To this end, two sets of experiments were conducted. Firstly, a baseline that fine-tunes the language models on predicting the hard label is defined, and then two architectures for incorporating the judgment of the annotator demographic profile were considered as the intervention models. The following items describe these three architectures in more detail and Figure 1 offers a visual representation of the design of the multi-task learning systems:

\begin{itemize}
    \item \textbf{STL-full-FT (baseline)}: This approach follows a standard single-task learning architecture that simply updates the language model parameters to adapt its knowledge to the task at hand. This architecture can be seen as a straightforward and popular choice among the NLP community and practitioners \cite{devlin2018bert, howard2018universal}.
    \item \textbf{MTL-six-aux}: The first multi-task learning architecture being introduced uses the hard label as the main task and incorporates the six additional labels from annotators as auxiliary tasks. These labels aligned with annotator profiles which were previously defined based on their demographic traits of gender and age group.
    \item \textbf{MTL-two-aux}: Here, the hard label served as the main task as well but the architecture performed an extra step to reduce the six annotator profiles to two. Specifically, the responses from MALE and FEMALE annotators were aggregated as two labels and were utilized as the auxiliary tasks of the multi-task system. This architecture is essentially different because it represents the collective perspective of all male and female annotators in a more condensed way that aligns with the expected polarity of the phenomenon in practice.
\end{itemize}

\begin{figure*}[t]
\label{f2}
\includegraphics[width=12cm]{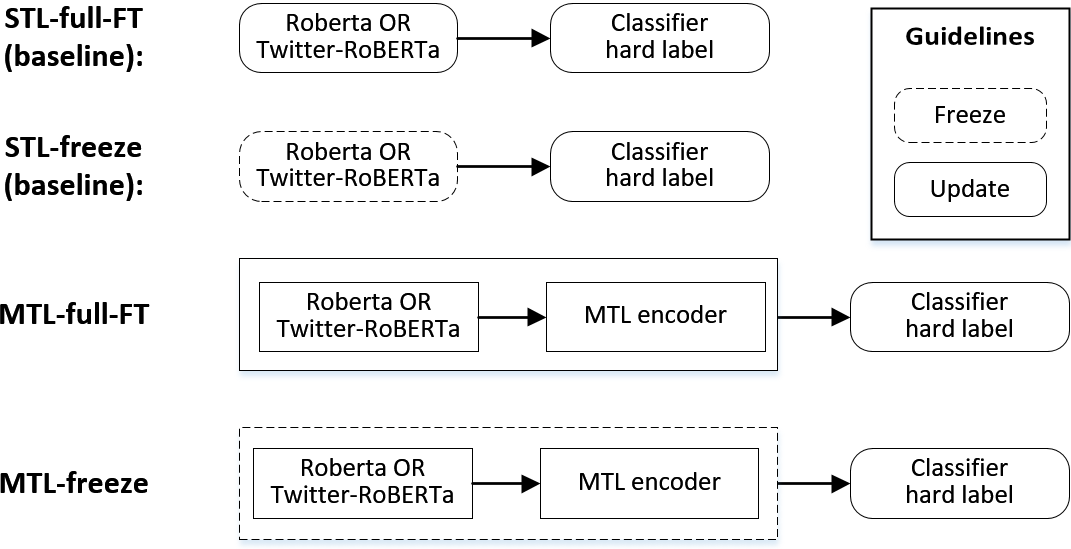}
\caption{Comparison of Single-task learning and Multi-task learning architectures when applying the variation of freezing versus updating the model parameters}
\end{figure*}

Subsequently, a second set of experiments was designed to explore the impact of parameter freezing or updating within the architecture on the system's adaptation to the task. This investigation provides valuable insights into how the nature of the acquired knowledge depends on the chosen approach in addressing the task. The following items explain the architectural variations while Figure 2 offers a visual representation to describe the layers of the architecture being frozen or fine-tuned:

\begin{itemize}
    \item \textbf{STL-full-FT (baseline)}: This baseline is the same previously mentioned that simply updates the language model parameters to adapt its knowledge to the task at hand.
    \item \textbf{STL-freeze (baseline)}: This second baseline freezes the language model parameters. It can be interpreted as utilizing the raw language model knowledge without performing any adaptation for the current task.
    \item \textbf{MTL-full-FT}: This architecture incorporates an encoder layer that leverages the annotator profiles as auxiliary tasks following a multi-task learning approach and adds a second step of fine-tuning on the hard label while updating all model parameters in a nested fashion.
    \item \textbf{MTL-freeze}: Similar to MTL-full-FT but only trained the classifier head, keeping the backbone frozen.
\end{itemize}

To initialize these systems, we employed the RoBERTa-base and the Twitter-RoBERTa-base-2022 transformer-based models, both considered state-of-the-art for text classification. Our training of the language models was configured with the set of hyperparameters in the MaChAmp \cite{van-der-goot-etal-2021-massive} framework with minimal changes: maximum epoch of 10 and learning rate of $5e-5$.
We evaluated these models using the macro-F1 score as a performance metric. Due to the unavailability of the test split from the Exist2023 shared task, the train split with 3260 tweets was used for training and validation, while the original development split with 489 tweets served as our test set.



\section{Results}
Now, we present the important findings of our study, explaining how different models and methods perform in detecting misogynistic content as well as implications.

\begin{table*}[ht!]
\caption{Model performance of {Single--task} learning versus Multi-task learning considering two approaches for leveraging the perspectives of annotators }
\centering
\begin{tabular}{l ccc|ccc}
\hline
\multicolumn{1}{c}{} & \multicolumn{3}{c|}{RoBERTa} & \multicolumn{3}{c}{Twitter-RoBERTa}
\\
\multicolumn{1}{c}{Architecture} 
& \multicolumn{1}{c}{Precision} & \multicolumn{1}{c}{Recall} & \multicolumn{1}{c|}{F1-score}
& \multicolumn{1}{c}{Precision} & \multicolumn{1}{c}{Recall} & \multicolumn{1}{c}{F1-score}
\\ \hline

STL-full-FT (baseline) & 0.842 & \textbf{0.843} & 0.842 & 0.831 & 0.820 & 0.824 \\
MTL-six-aux & 0.830 & 0.831 & 0.831 & 0.856 & 0.862 & 0.857 \\
MTL-two-aux & \textbf{0.859} & 0.837 & \textbf{0.843} & \textbf{0.871} & \textbf{0.876} & \textbf{0.873} \\
\hline
\end{tabular}

\label{tab:result1}
\end{table*}

\begin{table*}[ht!]
\caption{Contrasting {Single--task} learning versus {Multi--task} learning models performance when freezing parameters}
\centering
\begin{tabular}{l ccc|ccc}
\hline
\multicolumn{1}{c}{} & \multicolumn{3}{c|}{RoBERTa} & \multicolumn{3}{c}{Twitter-RoBERTa}
\\
\multicolumn{1}{c}{Architecture} 
& \multicolumn{1}{c}{Precision} & \multicolumn{1}{c}{Recall} & \multicolumn{1}{c|}{F1-score}
& \multicolumn{1}{c}{Precision} & \multicolumn{1}{c}{Recall} & \multicolumn{1}{c}{F1-score}
\\ \hline

STL-full-FT (baseline) & 0.842 & 0.843 & 0.842 & 0.831 & 0.820 & 0.824 \\
STL-freeze (baseline) & 0.626 & 0.614 & 0.613 & 0.659 & 0.643 & 0.643 \\
MTL-six-full-FT & \textbf{0.857} & \textbf{0.854} & \textbf{0.855} & \textbf{0.860} & 0.847 & 0.851 \\
MTL-six-freeze & 0.840 & 0.806 & 0.812 & 0.856 & 0.855 & 0.856 \\
MTL-two-full-FT & 0.838 & 0.836 & 0.837 & 0.844 & 0.850 & 0.844 \\
MTL-two-freeze & 0.846 & 0.841 & 0.843 & \textbf{0.862} & \textbf{0.863} & \textbf{0.863} \\
\hline
\end{tabular}

\label{tab:result2}
\end{table*}

 Our comprehensive analysis encompasses a range of approaches, including single-task learning (STL) and multi-task learning (MTL) architectures. Specifically, we conducted two sets of experiments. 

The first set of experiments investigates how the method for incorporating the annotators' judgment affects the learning capability of the systems when using the RoBERTa--base and the Twitter--RoBERTa-base-2022 transformer-based models as initialization. In particular, the baseline system only finetunes the language model on the hard label which corresponds to the majority vote of all annotators. Our MTL architectures incorporate the annotator perspectives as auxiliary tasks that correspond to different segmentations of their demographic profile. Table \ref{tab:result1} summarizes the results we obtained from these experiments.

The second part of our experiments offered valuable insights into the efficacy of the proposed single-task and multi-task methodologies when updating or freezing, which is a popular choice to trade-off the system learning capabilities with the time it takes to train. This can guide future research and the development of practical applications aimed at combating online misogyny. The experiments included both approaches for incoporating the annotators' perspectives, namely the MTL-six-aux and MTL-two-aux archictures. Table \ref{tab:result2} summarizes the results we obtained from these experiments.

Our study underscores the effectiveness of the multitask learning (MTL) approach in the context of misogynistic content detection. Across both the RoBERTa and Twitter-RoBERTa language models all MTL models consistently performed comparably or outperformed the single-task learning (STL) models, offering strong evidence of the advantages of incorporating multiple perspectives as auxiliary tasks. However, it's worth noting that one exception was observed: the MTL-freeze model showed a more evident lower performance when using the RoBERTa model. This suggests that MTL is generally a robust strategy but may require careful consideration of specific model configurations.

When it comes to the number of profile groups employed, our study found that there isn't a substantial difference in performance between using six profile groups and using two profile groups. However, the MTL model aggregating all six profiles into MALES versus FEMALES tends to perform slightly better, hinting at the potential advantages of considering the nuances of partially aggregating perspectives when modeling the subjectivity of annotators as auxiliary tasks.

In our examination of whether to freeze or update language model parameters, we observed that, as anticipated, freezing parameters can negatively impact model performance. This effect is particularly pronounced in the single-task learning model (the baseline). However, the proposed multi-task learning model exhibited a notably robust behavior. What's more, in the case of the MTL-two-freeze models, they outperformed their fine-tuning counterparts (MTL-two-full-FT) without the need for the nested parameter updating process.

Moreover, our research highlights a trade-off between performance and simplicity within the MTL models. While the results from the models using the MTL-six-full-FT and MTL-two-full-FT architectures exhibited good performance across both language models, the models using MTL-six-aux and MTL-two-aux provided comparable results while being simple as they do not include a nested finetuning step in their architecture. Depending on the specific requirements and constraints of the application, the archictures incorporating the second funetunning step may emerge as a preferable option for implementation.

Lastly, our investigation into the choice of language models unveiled that Twitter-RoBERTa-base-2022 showcased a slight performance edge compared to the RoBERTa-base model. This finding suggests that the choice of language model can influence the system's efficacy in identifying misogynistic content. These conclusions provide relevant knowledge for researchers and practitioners in the field, guiding future studies and applications in combating online misogyny.

\section{Error analysis}

\begin{table*}[t]
    \caption{Samples from challenging cases expressing indirect misogyny}
    \centering
    \begin{tabular}{p{0.07\linewidth} p{0.26\linewidth}  p{0.26\linewidth}  p{0.26\linewidth}}
    \hline
        Error type & Claim about misogyny & Complex references & No enough context \\
    \hline
        False \newline positive &
        Ladies, don't have a miscarriage in Louisiana! If you do, in addition to probably having your friends and family sued by some opportunistic yokel, you'll be charged with murder. url                                                      
         &
         @user My new favorite meme is the pregnant woman driving in the Oklahoma HOV lane telling the state trooper who pulled her over that she is driving with another person..
         &  
         @user Exactly how I feel. I’m gonna be alone for sure,  it’s almost sexist / racist / whatever else ist just to ask this anymore
         \\

         False negative &
         @user Have I missed something, or has phallocentrism become the latest cult?
         &
         @user Best way to do that is from behind her, with your hands on her hips
         &
         @user yea back when i didn’t even have one to slap, unfortunate !
         \\
    \hline
    \end{tabular}
    \label{tab:error-analysis}
\end{table*}

For our error analysis, we selected four models to represent our study: the baseline STL-full-FT and the MTL-six-aux architectures, each using the RoBERTa and Twitter-RoBERTa models.

Our analysis focused on instances where all four models failed to align with the hard label, which was determined by a majority vote from six annotators. We considered these cases to be among the most challenging for classification.

Upon examination, we observed that these samples often involved the indirect expression of misogyny. In these instances, misogyny was not explicitly stated but could be inferred or indirectly implied through contextual cues, tone, or underlying messages in the content. Analyzing such scenarios often necessitates a more in-depth understanding of contextual nuances to identify subtle forms of discrimination or harmful attitudes toward women.

The following items outline some possible reasons for the systems' misclassification in these challenging cases, categorized into three areas. A concrete list of samples for this categorization is offered in Table \ref{tab:error-analysis}:
\begin{itemize}
    \item  \textbf{Claim about misogyny:} Cases where the text is possibly stating a position or a call to action about misogyny	
    \item  \textbf{Complex references:} Narrative of scenarios where references are being used and may be hard to relate to the expression of misogyny	
    \item \textbf{No enough context:} Cases in which the provided context is probably not enough for a system to make a justified prediction
\end{itemize}

In examining the false positive cases, we observed instances where the models erroneously classified content as implying misogyny. This misclassification probably occurred due to the presence of related terms or phrases, such as 'miscarriage', which triggered the model, or the misinterpretation of references that were not directly related to misogyny, for example, 'pregnant woman driving'. Additionally, some false positives might have resulted from situations where the model lacked sufficient context to make a well-informed decision as shown in Table \ref{tab:error-analysis}. Conversely, in the false negative cases, we found occasions where the model might have failed to recognize text discussing misogyny. For instance, phrases like 'the cult of phallocentrism' were not identified as misogynistic content. Additionally, the model could not have been able to connect indirect references to misogyny due to its limitations in spatial knowledge or reasoning, or the absence of contextual information associated with insulting women. These challenges in identifying indirect expressions of misogyny within textual data can be considered some of the most difficult obstacles to achieving accurate classifications by the model.

\section{Conclusion}
Our study contributed to advancing the understanding of the complex task of misogynistic content detection in online spaces. We have unveiled critical insights by examining different model architectures, profile group considerations, and language model choices. Multitask learning (MTL) emerges as a powerful approach, consistently outperforming single-task learning (STL) in our evaluations. It underscores the importance of incorporating diverse perspectives from annotators of various demographic backgrounds in the fight against online misogyny. Additionally, our exploration into language model selection demonstrates that nuanced performance differences exist, with RoBERTa-twitter exhibiting a slight edge over the base RoBERTa model. These findings collectively contribute to the ongoing effort to create more effective content moderation systems, emphasizing the significance of MTL and the careful consideration of profile groups in addressing this critical societal challenge.

As we move forward, our research paves the way for continued investigations into mitigating biases and subjectivity in content moderation systems, encouraging a more inclusive and equitable online environment for all users. Finally, the insights gained from this study can help inform the development of robust and effective tools to combat the pervasive issue of online misogyny, promoting safer and more respectful online interactions.

\section{Future work}
The advances in the systems able to identify misogyny can help victims of online abuse receive quicker support, and platforms can take action more promptly and therefore discourage potential perpetrators. 
With our findings in mind, we see great potential in further investigating subjectivity within misogynistic texts, especially exploring how cultural backgrounds and exposure to various forms of misogyny impact interpretations, an area that has been insufficiently explored. As noted in \cite{annotatingmisogyny2021}, expressions of misogyny notably vary based on the demographic profile of the audience. For instance, Spanish expressions tend to emphasize "Dominance," Italian expressions often feature "stereotyping and objectification," English discussions often involve "discrediting" women, and Danish discussions predominantly touch upon "neo-sexism," a viewpoint that denies the existence of misogyny in contemporary times. In our future research, we aim to shed light on the potential biases and subjective judgments that can arise when annotators from diverse cultural backgrounds assess whether a given text qualifies as misogynistic or not.

Then, additional research is required to assess our findings in different languages, sources of information, and demographic groups whose contexts and cultural criteria may differ considerably from the beliefs expressed in the annotations of this study. In the present study, we only made experiments for English, and more specifically the text being used was extracted from the Twitter platform (now called X) from specific demographic profile groups.

Lastly, we intend to explore additional datasets containing demographic information of annotators and strongly encourage dataset creators to provide this valuable information, as demonstrated by its utility in our research.


\section*{Acknowledgements}
The work was done with partial support from the Mexican Government through the grant A1-S-47854 of CONACYT, Mexico, grants 20232138, 20232080, 20231567 of the Secretaría de Investigación y Posgrado of the Instituto Politécnico Nacional, Mexico. The authors thank the CONACYT for the computing resources brought to them through the Plataforma de Aprendizaje Profundo para Tecnologías del Lenguaje of the Laboratorio de Supercómputo of the INAOE, Mexico and acknowledge the support of Microsoft through the Microsoft Latin America PhD Award.

\bibliography{custom}
\bibliographystyle{abbrv}







\end{document}